\documentclass{article}

\usepackage{PRIMEarxiv}

\usepackage[utf8]{inputenc} 
\usepackage[T1]{fontenc}    
\usepackage{hyperref}       
\usepackage{url}            
\usepackage{booktabs}       
\usepackage{amsfonts}       
\usepackage{nicefrac}       
\usepackage{microtype}      
\usepackage{lipsum}
\usepackage{fancyhdr}       
\usepackage{graphicx}       
\graphicspath{{media/}}     

\usepackage{setspace}

\usepackage{amsmath}
\usepackage{amssymb}

\usepackage{algorithm}
\usepackage{algpseudocode}
\usepackage{algorithmicx}

\usepackage{threeparttable}    
\usepackage{multirow}

\title{Auxiliary-Tasks Learning for Physics-Informed Neural Network-Based Partial Differential Equations Solving \thanks{The manuscript is under review.}
}

\author{
  Junjun Yan$^1$, Xinhai Chen$^{1,}$\thanks{Corresponding author} , Zhichao Wang$^1$, Enqiang Zhou$^2$ and Jie Liu$^2$ \\
  $^1$ Science and Technology on Parallel and Distributed Processing Laboratory,\\
  National University of Defense Technology, Changsha, 410073, China\\
  $^2$ Laboratory of Digitizing Software for Frontier Equipment,\\ 
  National University of Defense Technology, Changsha, 410073, China\\
  \texttt{\{yanjunjun, chenxinhai16$^*$\}@nudt.edu.cn} 
}

\begin{document}
\maketitle

\begin{abstract}
Physics-informed neural networks (PINNs) have emerged as promising surrogate modes for solving partial differential equations (PDEs). Their effectiveness lies in the ability to capture solution-related features through neural networks. However, original PINNs often suffer from bottlenecks, such as low accuracy and non-convergence, limiting their applicability in complex physical contexts. To alleviate these issues, we proposed auxiliary-task learning-based physics-informed neural networks (ATL-PINNs), which provide four different auxiliary-task learning modes and investigate their performance compared with original PINNs. We also employ the gradient cosine similarity algorithm to integrate auxiliary problem loss with the primary problem loss in ATL-PINNs, which aims to enhance the effectiveness of the auxiliary-task learning modes. To the best of our knowledge, this is the first study to introduce auxiliary-task learning modes in the context of physics-informed learning. We conduct experiments on three PDE problems across different fields and scenarios. Our findings demonstrate that the proposed auxiliary-task learning modes can significantly improve solution accuracy, achieving a maximum performance boost of 96.62\% (averaging 28.23\%) compared to the original single-task PINNs. The code and dataset are open source at https://github.com/junjun-yan/ATL-PINN.
\end{abstract}


\section{Introduction}

Partial differential equations (PDEs) play a crucial role in describing numerous essential problems in physics and engineering fields. However, numerical iterative solvers can be computationally expensive when solving inverse problems, high-dimensional problems, and problems involving complex geometries, despite the accurate approximation provided by numerical solvers \cite{ns1, HFM, nd1, phy-inf}. The development of artificial intelligence methods, typically deep neural networks (DNNs), has facilitated the successful application of data-driven models in such fields. The DNNs models effectively reduce computational time and enhance efficiency. Physics-informed neural networks (PINNs), a class of DNNs that incorporate PDEs into the loss function, transform the numerical problem into an optimization problem. In the early 1990s, research efforts had been devoted to solving PDEs using neural networks \cite{old-phy}. However, due to the limitations in hardware and algorithm, this method did not receive significant attention. In recent years, the introduction of the PINN training architecture by Raissi et al. \cite{PINN} sparked interest in physics-informed learning within this interdisciplinary field.

Compared to the traditional numerical solvers, PINNs offer several advantages \cite{pinnrev, app4, app5}. Firstly, PINNs can handle inverse problems as straightforward as forward problems by treating the unknown parameters as trainable variables and learning them through back-propagation during network training. Secondly, PINNs alleviate the issue of exponential growth in network weight size with increasing dimensions, effectively bypassing the dimension catastrophe commonly encountered in traditional methods. Another significant advantage of PINNs is their mesh-free nature. They can directly sample points in complex geometric domains for training without computationally-expensive mesh generation. Therefore, PINNs possess the potential to overcome the challenges faced by traditional methods. Despite some studies demonstrating the convergence of PINNs in several scenarios, their accuracy remains inadequate \cite{theory2, shin, mishra}. PINNs are hard to train due to the statistical inefficiency of building useful representations from physics information, thereby leaving room for enhancing the performance by improving the representations in the latent solution space \cite{theory1}.

Auxiliary-task learning, a prevalent technique in deep learning applications such as computer vision, natural language processing, and recommender systems, falls under the umbrella of multi-task learning \cite{mt_review1, AT1, AT2, AT3}. In this framework, the primary problem referred to as the main task, is accompanied by similar problems known as auxiliary tasks, which can potentially enhance the performance of the main task. The fundamental concept behind auxiliary-task learning involves leveraging a shared representation to learn from both tasks. When the auxiliary tasks are closely related to the main task, this approach will improve prediction accuracy and reduces data requirements \cite{AT_review}. In solving PDEs, tasks involving the same governing equation but different boundaries or initial conditions can be treated as correlated tasks. Despite the potential heterogeneity of the final physics fields, there is still shared physical information among them. Consequently, by fully capitalizing on auxiliary tasks through a shared representation, we can potentially enhance the accuracy of the main tasks. 

This paper presents the pioneering research that integrates an auxiliary-task learning mechanism into physics-informed learning. Specifically, we implement four distinct network structures for auxiliary-task learning within a physics-informed framework. Furthermore, we introduce the gradient cosine similarity algorithm to all auxiliary-task networks, ensuring that the main task consistently benefits from the auxiliary task in terms of gradients \cite{GCS}. To validate our approach, we conduct a comprehensive set of experiments using three different PDEs from the PDEBench dataset \cite{pdebench}. The experimental results demonstrate that auxiliary-task learning can effectively enhance the DNN-based physics-informed surrogate models. Overall, by actively constructing auxiliary tasks through variations in the initial conditions of the same PDEs, the prediction accuracy is improved by an average of 28.23\% and a maximum of 96.62\% across the experimental datasets. These findings highlight the general utility of auxiliary-task learning as a technique for enhancing the performance of surrogate models.

The remainder paper is organized as follows: Section \ref{Background} provides an overview of the related work and background on PINNs, multi-task learning, and auxiliary-task learning. Section \ref{method} presents detailed descriptions of four auxiliary-task learning network structures, along with an introduction to the gradient cosine similarity algorithm. Section \ref{experiments} shows the numerical experiments conducted to evaluate the performance of different auxiliary-task networks across various PDEs. Finally, Section \ref{Conclusion} concludes the paper and discusses future directions for research.

\section{Background}
\label{Background}

We first define the initial-boundary value problem of general PDEs as follows:
\begin{gather}
N\left[u(x,t);\lambda\right]=f(x,t), x\in\Omega, t\in(0,T] \label{eq1} \\
B[u(x,t)]=g(x,t), x\in\partial\Omega, t\in(0,T] \label{eq2} \\
u\left(x,0\right)=h(x), x\in\bar{\Omega} \label{eq3}
\end{gather}
Here, $\lambda$ is the unknown equation parameters; $u(x,t)$ denotes the latent (hidden) solution at time $t$ and location $x$; $f(x,t)$ is the equation forcing function; $g(x,t)$ and $h(x)$ are boundary conditions function and initial conditions function respectively; $N[\cdot]$ and $B(\cdot)$ are nonlinear differential operators, where $B(\cdot)$ can be Dirichlet, Neumann, or mixed boundary conditions. We note that $\Omega\subset\mathbb{R}^d$ in an open set while $\bar{\Omega}$ is its closure.

\subsection{Physics-informed neural networks}

\begin{figure}[htbp]
  \centering
  \includegraphics[width=0.75\linewidth]{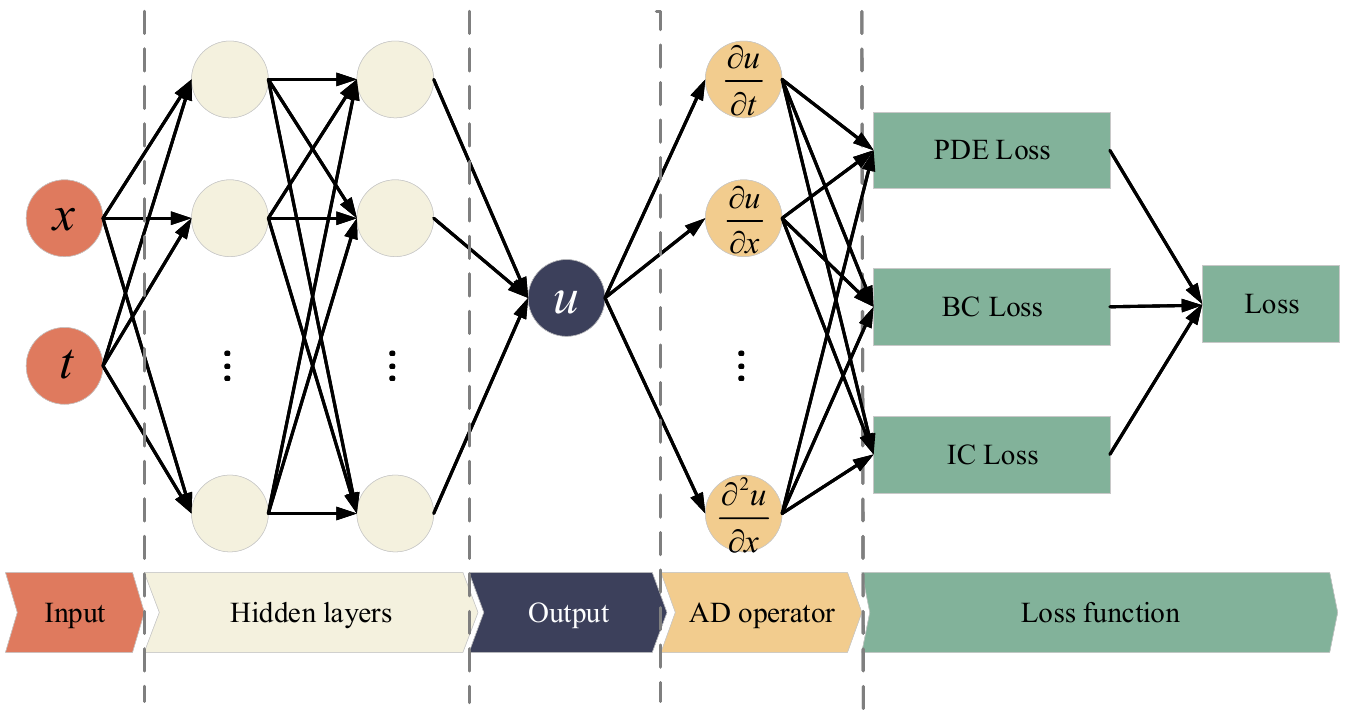}
  \caption{An example of a physics-informed neural network.}
  \label{fig1}
\end{figure}

PINN is a surrogate model based on deep neural networks that utilize temporal-spatial coordinates $(x, t)$ as input and predicts the physics field $(u)$ as output. In contrast to traditional fully connected neural networks, PINNs incorporate governing equations, boundary conditions, and initial conditions into the loss function, which ensures that the network output adheres to these constraints. Consequently, PINNs not only learn from the data distribution but also conform to the laws of physics. The loss function comprises several components: \eqref{eq4} represents the loss function of the governing equations, \eqref{eq5} denotes the loss function of the boundary conditions, and \eqref{eq6} indicates the loss function for the initial conditions. While PINNs can train the network without supervised data, practical applications often include limited data points within the domain (sensor data) to expedite model convergence. The loss function for these data points is represented by \eqref{eq7}. Finally, all the loss components are weighted and summed together as \eqref{eq8}, allowing for training through backpropagation and gradient descent.
\begin{gather}
L_f=\frac{1}{N_f}\sum_{i=1}^{N_f}\left|N\left[\hat{u}(x_f^i,t_f^i);\lambda\right]-f(x_f^i,t_f^i)\right|^2 \label{eq4} \\
L_b=\frac{1}{N_b}\sum_{i=1}^{N_b}\left|B\left[\hat{u}(x_b^i,t_b^i)\right]-g(x_b^i,t_b^i)\right|^2 \label{eq5} \\
L_0=\frac{1}{N_0}\sum_{i=1}^{N_0}\left|\hat{u}(x_0^i,\ 0)-h(x_0^i)\right|^2 \label{eq6} \\
L_p=\frac{1}{N_p}\sum_{i=1}^{N_p}\left|\hat{u}(x_p^i,\ t_p^i)-u_p^i\right|^2 \label{eq7} \\
L={W_f\ast L}_f+W_b\ast L_b+{W_0\ast L}_0+{W_d\ast L}_d \label{eq8}
\end{gather}
where $(x_f^i,t_f^i)$, $(x_b^i,t_b^i)$, $(x_0^i,0)$ and $(x_d^i,t_d^i)$ represent the sample points, boundary points, initial points, and data points (if available). The predicted solution by the network, used to approximate $u$, is denoted as $\hat{u}$, $y_d^i$ represents the supervised real solution (if available). The weights assigned to each loss component are $W_f$, $W_b$, $W_0$, and $W_d$, respectively. Figure \ref{fig1} illustrates the architecture of PINN, and the deep learning frameworks (e.g., TensorFlow, PyTorch) enable convenient learning of high-order gradient losses through their automatic differentiation mechanisms (AD).

The baseline PINN method proposed by Raissi et al. \cite{PINN} achieved tremendous success in solving PDE problems in many physics and engineering fields, such as fluid mechanics, quantum mechanics, and electromagnetism \cite{app1, app2, app3}. Several articles analyzed that the PINNs can converge to the solution in some satiations. For example, Shin et. al. \cite{shin} first demonstrated that in the Holder continuity assumptions, the upper bound on the generalization error is controlled by training error and the number of training data. At the same time, Mishra et. al. \cite{mishra} establish a theory of using PINNs to solve forward PDE problems. They get a similar error estimation result under a weaker assumption and generalize the result to the inverse problem \cite{theory2}. However, in practice cases, the PINNs sometimes fail to train. Wang et. al. \cite{theory1} analyzed this phenomenon from a neural tangent kernel (NTK) perspective. They find the convergence rate of different parts in the loss function has a remarkable discrepancy and they suggested utilizing the eigenvalues of the NTK matrix to adaptively the weight of loss to solve the above problems.

A substantial amount of prior research has been dedicated to enhancing the prediction accuracy and training effectiveness of the baseline PINNs by introducing novel network architectures and training methods. For instance, Wu et al. \cite{adp} proposed the residual-based adaptive refinement method, which strategically samples additional points in high-loss regions. Yu et al. \cite{gpinn} introduced g-PINN, which incorporates higher gradient information into the loss function. Recognizing the significance of weights in PINN training, McClenny et al. \cite{SAPINN} developed a self-attention PINN (SA-PINN) that adapts training weights for each data point using a soft multiplicative attention mask mechanism similar to those used in computer vision. Differing from SA-PINN, competitive physics-informed networks, introduced by Zeng et al. \cite{COM_PINN}, employ a discriminator to reward the prediction of errors made by the PINN. Several papers have explored the domain decomposition approach \cite{cpinn, xpinn}. As PINN training aligns with semi-supervised learning scenarios, Yan et al. \cite{stpinn} incorporated a self-training mechanism into PINN training, utilizing the residual of the physics equation as an index for generating pseudo-labels. In addition, there are many practical problems that PINNs have been successfully applied to, including fluid mechanics, mechanics of materials, power systems, and biomedical \cite{cxh1, cxh2, PINN-FFHT}. While the above studies cover a broad spectrum of research on physics-informed learning, few papers have explored the application of auxiliary-task learning methods to PINNs.

\subsection{Auxiliary-tasks learning}

Multi-task learning (MTL) is a machine learning approach that can enhance learning performance and efficacy by utilizing a shared representation for multiple tasks. Fine-tuning is a specific instance of multi-task learning as it leverages different tasks sequentially. In contrast to learning individually, multi-task learning potentially improves overall performance. In recent years, multi-task learning has emerged as a powerful approach for enhancing model training performance and effectiveness, finding successful applications in various domains of deep learning, including computer vision, natural language processing, and recommendation systems \cite{mt1, mt_review1, MMoE, PLE}.

Baxter et al. \cite{mt2} suggest that MTL improves generalization performance by incorporating domain knowledge learned from the supervised signal of relevant tasks. When a hypothesis space performs well on a sufficient number of training tasks, it is more likely to perform well on new tasks within the same environment, thus facilitating generalization. Furthermore, MTL enables the model to leverage knowledge from other tasks, allowing for learning features that a single task alone cannot capture \cite{mt4}. Ruder et al. \cite{mt3} provides a biological interpretation of this phenomenon, drawing parallels to how infants learn to recognize faces and utilize this knowledge to recognize other objects later. The authors argue that MTL better captures the learning process observed in human intelligence, as integrating knowledge across domains is a fundamental aspect of human cognition.

The multi-task learning approach can be divided into two classes based on the shared mode of hidden-layer parameters: hard-shared and soft-shared. The hard sharing mechanism is the most commonly used approach, which can date back to the literature \cite{mt5}. Generally, it applies to all hidden layers for all tasks while leaving the task-specific output layers. This mechanism reduces the risk of overfitting. However, the hard sharing mode is sensitive to the similarity between different tasks, potentially leading to interference among non-similar tasks. As for the soft-shared, each task has its own parameters. The similarity of parameters is ensured through regularization techniques. For instance, \cite{mt6} uses L2 distance regularization, while \cite{mt7} employs trace norm regularization. The soft-shared mechanisms in deep neural networks draw inspiration from regularization techniques in traditional multi-task learning and can mitigate the challenges faced by hard sharing. Consequently, soft sharing has emerged as a focal point of modern research.

Auxiliary-task learning is a specific form of multi-task learning that shares a common theoretical foundation but diverges in its implementation. While traditional multi-task learning aims to enhance the performance of all tasks within a single framework, auxiliary-task learning focuses on utilizing additional tasks to improve the performance of the primary task independently. Recent studies have underscored the importance of fully leveraging auxiliary tasks to enhance the main task's performance, leading to consistent benefits \cite{AT1,AT2,AT3,AT_review, GCS}. However, limited research has been conducted in the context of physics-informed learning problems. In solving PDEs, one governing equation with two different initial conditions can be considered two distinct tasks. They may address two different hypothesis spaces while connecting through the shared governing equation. Consequently, it is promising for PINNs to achieve more accurate predictions by leveraging knowledge from other tasks. In this study, we explore four distinct network structures for auxiliary-task learning and evaluate their performance on three partial differential equation problems, which we describe in the subsequent sections.

\section{Auxiliary-task learning-based physics-informed neural networks}
\label{method}

To investigate the potential improvement of PINNs through auxiliary-task learning, we conducted experiments involving different PDEs and network architectures. In practical problems, tasks can be created by randomly altering the initial condition. However, we directly selected diverse tasks from the PDEBench dataset during our experiments, which is more convenient. The PDEBench dataset \cite{pdebench} is a novel resource for scientific machine learning. It provides a diverse range of PDE problems with varying initial conditions, which create conditions for our experiments. In Section \ref{baseline}, we will present the network architecture for auxiliary-task learning as implemented in our paper. Furthermore, in Section \ref{GCS}, we will introduce the gradient cosine similarity algorithm, which optimizes the above models from the perspective of a gradient in physics-informed learning.

\subsection{The proposed ATL-PINN modes}
\label{baseline}

We provide four auxiliary-task learning modes below, which will be compared with the original PINNs to demonstrate the potential impact of auxiliary-task learning in physics-informed learning.

\begin{figure}[htbp]
  \centering
  \includegraphics[width=0.9\linewidth]{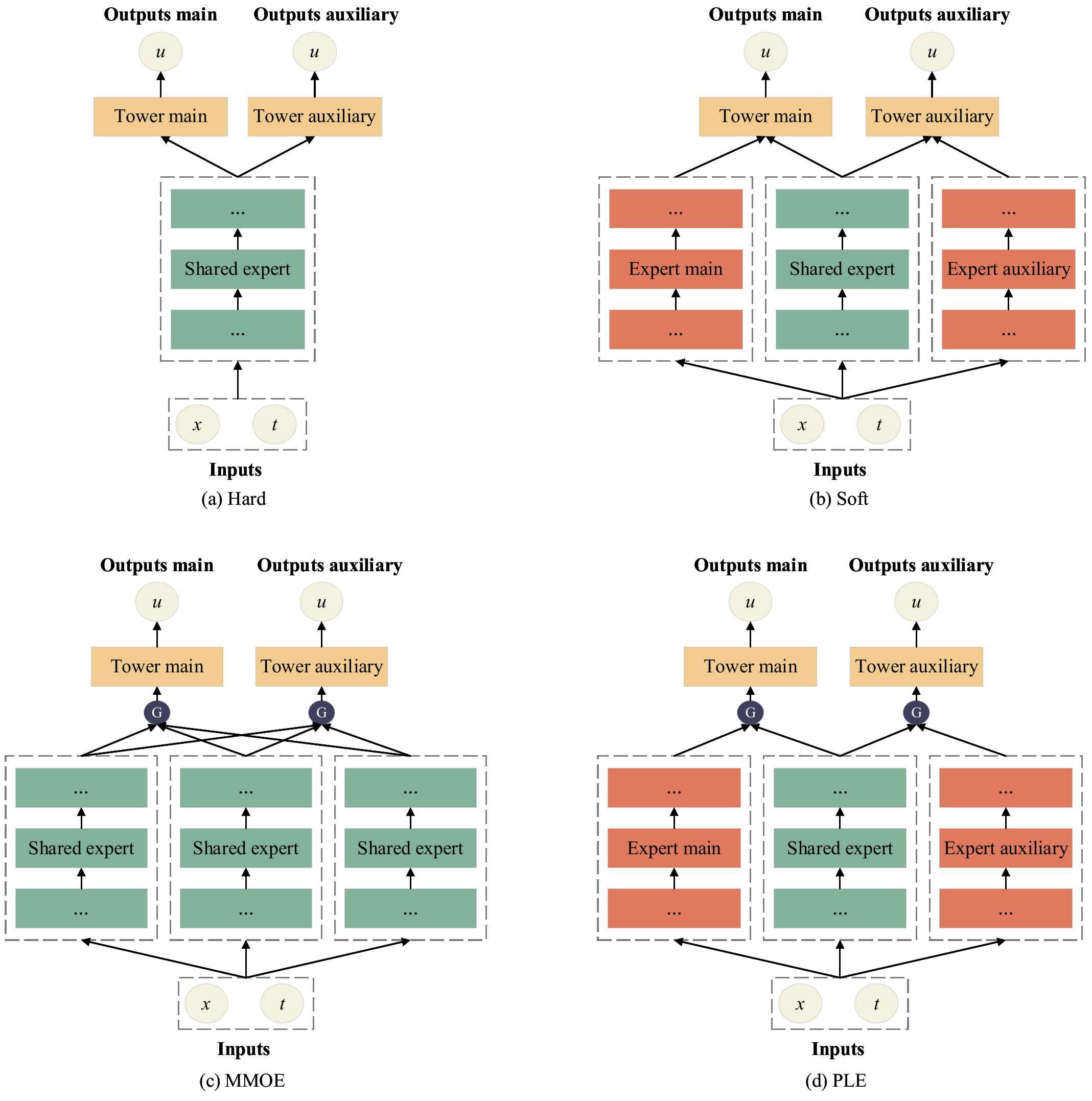}
  \caption{The architecture sketch of four different auxiliary-task learning modes.}
  \label{fig2}
\end{figure}

\textbf{Hard-shared auxiliary-task learning-based physics-informed neural network (Hard-ATL-PINN)}. Figure \ref{fig2} (a) depicts the architecture of the hard-shared network. In this approach, the main and auxiliary tasks share an expert for feature extraction. After that, each task has a non-shared specific tower network to perform the final processing. The hard sharing mechanism is a widely employed strategy in multi-task learning of neural networks, known for its effectiveness in various problems.

\textbf{Soft-shared auxiliary-task learning-based physics-informed neural network (Soft-ATL-PINN)}. The network architecture depicted in Figure \ref{fig2} (b) represents a customized version of a soft-shared network. It comprises shared expert networks and task-specific expert networks for each task. The underlying concept of this network design is to separate shared and specific features by training experts on different data. The network structure is flexible, which allows us to custom-define the number of shared / task-specific experts. 

\textbf{Multi-gate mixture-of-experts auxiliary-task learning-based physics-informed neural network (MMoE-ATL-PINN)}. The feature extraction network layer consists of multiple expert networks shared by multiple tasks and a single gate network unique to each task, known as the Multi-gate Mixture-of-Experts (MMoE) layer (shown in Figure \ref{fig2} (c)). The MMoE layer is designed to enable the modes to automatically learn how to allocate experts based on the relationships among the underlying tasks. It introduces the gate network as a self-attention mechanism, which has been successfully applied in various backbones. In scenarios where the underlying task relationship is weak, the model can learn to activate only one expert per task, effectively assigning different experts to different tasks.

\textbf{Progressive layered extraction auxiliary-task learning-based physics-informed neural network (PLE-ATL-PINN)}. Compared to the MMoE modes, which employs only shared expert networks, PLE incorporates both shared and task-specific expert networks (shown in Figure \ref{fig2} (d)). This design can guide the mode to learn the common feature through shared networks and the private feature through task-specific expert networks, thereby alleviating the seesaw phenomenon when the relationships between tasks are weak. The PLE method divides the mode's parameter representation into private and public parts for each task, enhancing the robustness of auxiliary-task learning and mitigating negative interactions between task-specific pieces of knowledge. 

\subsection{Gradient cosine similarity algorithm for auxiliary-task learning}
\label{GCS}

We employ gradient cosine similarity to leverage the auxiliary loss in conjunction with the main loss. The gradient cosine similarity is defined as follows: 
\begin{equation}
    \mathrm{Cosine\ Similarity}\left(\theta\right)=\frac{\nabla_\theta\mathcal{L}_{main}\left(\theta\right)\cdot\nabla_\theta\mathcal{L}_{aux}\left(\theta\right)}{|\nabla_\theta\mathcal{L}_{main}\left(\theta\right)||\nabla_\theta\mathcal{L}_{aux}\left(\theta\right)|}
    \label{eq9}
\end{equation}
where $\theta$ represents the shared network parameters, $\mathcal{L}_{main}\left(\theta\right)$ denotes the loss function of the main task, and $\mathcal{L}_{aux}\left(\theta\right)$ is the loss function associated with the auxiliary task. The gradient cosine similarity measures the degree of correlation between the tasks, thereby approximating the extent to which the gradient descent directions align between the main task and the auxiliary tasks.

If the gradient cosine similarity between the main task and the auxiliary task is positive, the network will update both their gradients (indicating that they share the same gradient direction). Conversely, if the gradient cosine similarity is negative, the network only updates the gradient for the main task and ignores the auxiliary task (indicating an adverse gradient direction). This update ensures the appropriate adjustment of the network's shared parameters, following Algorithm \ref{algroithm:1}. The symbol $\theta$ denotes the shared parameters, $\phi_{\mathrm{main}}$ and $\phi_{\mathrm{aux}}$ correspond to the private parameters of the main and auxiliary tasks, respectively. The symbol $\alpha$ represents the learning rate used in the training process. $\mathcal{L}_{main}$ and $\mathcal{L}_{aux}$ denotes the loss of the main task and auxiliary task, respectively. In this paper, we assess whether the cosine similarity between task gradients can serve as a reliable signal for identifying when the incorporation of an auxiliary loss benefits the main loss in physics-informed auxiliary task learning. To this end, we compare the non-cosine version with the cosine version of the approach.

\begin{algorithm}[h]
    \setstretch{1.35}
    \renewcommand{\algorithmicrequire}{\textbf{Input:}}
    \renewcommand{\algorithmicensure}{\textbf{Output:}}
    \caption{Gradient descent based on gradient cosine similarity}
    \begin{algorithmic}[1]
        \Require $\theta^{\left(t\right)},\phi_{main}^{\left(t\right)},\phi_{aux}^{\left(t\right)},\alpha^{\left(t\right)}$
        \Ensure $\theta^{\left(t+1\right)},\phi_{main}^{\left(t+1\right)},\phi_{aux}^{\left(t+1\right)}$
        \State Compute $\mathcal{L}_{main}$, $\mathcal{L}_{aux}$
        \State Compute $\nabla_{\phi_{main}}\mathcal{L}_{main}$, $\nabla_\theta\mathcal{L}_{main}$, $\nabla_{\phi_{aux}}\mathcal{L}_{aux}$, $\nabla_\theta\mathcal{L}_{aux}$
        \State $\phi_{main}^{\left(t+1\right)}\gets\phi_{main}^{\left(t\right)}-\alpha^{\left(t\right)}\nabla_{\phi_{main}}\mathcal{L}_{main}\left(\theta,\phi_{main}\right)$
        \State $\phi_{aux}^{\left(t+1\right)}\gets\phi_{aux}^{\left(t\right)}-\alpha^{\left(t\right)}\nabla_{\phi_{aux}}\mathcal{L}_{aux}\left(\theta,\phi_{aux}\right)$
        \If{$\mathrm{Cosine\ Similarity}(\nabla_\theta\mathcal{L}_{main}(\theta),\nabla_\theta\mathcal{L}_{aux}(\theta))>0$}
            \State $\theta^{\left(t+1\right)}\gets\theta^{\left(t\right)}-\alpha^{\left(t\right)}{(\nabla}_\theta\mathcal{L}_{main}\left(\theta\right)+\nabla_\theta\mathcal{L}_{aux}(\theta))$
        \Else
            \State $\theta^{\left(t+1\right)}\gets\theta^{\left(t\right)}-\alpha^{\left(t\right)}\nabla_\theta\mathcal{L}_{main}\left(\theta\right)$
        \EndIf
    \end{algorithmic}
    \label{algroithm:1}
\end{algorithm}

\section{Numerical experiments}
\label{experiments}

We conduct a series of experiments on three PDEs from various fields and levels of complexity. These include the one-dimensional time-space Diffusion Reaction equation, the one-dimensional time-space Burger’s equation, and the two-dimensional time-space Shallow Water equations. The training data used for these experiments are downloaded from the PDEBench dataset \cite{pdebench}, and the equations and parameters are set to their default values as specified in their paper. In all case study, the PEDBench dataset provides 10,000 tasks, each with a different initial condition. We randomly selected 100 tasks to create a subset dataset for auxiliary task learning. For each task, we randomly chose another task as its auxiliary task. Our deep learning framework of choice is Pytorch 1.12, and the experiments are performed using the NVIDIA Tesla P100 accelerator.

Table~\ref{tab1} displays the overall performance analysis. The modes are referred to as \emph{Org} for the original version and \emph{Cos} for the gradient cosine similarity version. \emph{L2 error} represents the average L2-related error across the 100 tasks under study; \emph{Boost} indicates the average percentage of improvement in prediction accuracy achieved by each mode; \emph{Number} quantifies the number of tasks in which the corresponding mode outperforms the PINN baseline. It is evident from the table that ATL-PINNs consistently outperform PINNs in terms of average errors. Notably, over 60\% tasks across all modes and PDE problems can benefit from auxiliary-task learning. The Diffusion Reaction equation exhibits the most significant improvement among ATL-PINN modes, achieving an average performance boost of 62.58\%. The Burgers' equation and the Shallow Water equation can be improved by approximately 14.30\% and 17.19\%, respectively. Despite the Burgers' equation showing the lowest average performance, it benefits the most expansive range of tasks, with nearly 90\% of problems in the training dataset exhibiting improvement. These results attest to the beneficial impact of learning in conjunction with auxiliary tasks in physics-informed learning.  

Regarding the gradient cosine similarity algorithm, notable improvements are observed in the Hard and Soft models across all three case studies compared to the original modes. The cosine similarity algorithm leads to an average enhancement of nearly 5\%. However, the benefits of MMoE and PLE modes are negligible. This discrepancy may be because of the attention mechanism introduced by the gate network. It allows the modes autonomously separate task-specific information into their respective experts. Therefore, the gate network performs efficiently regardless of using the cosine method. Notably, the Hard mode with the cosine method surpasses more complex modes such as MMoE and PLE, achieving the best results in the Shallow Water equation. In the subsequent sections, we will discuss the detailed experiment results for the three equations.  

\begin{table}[htbp]
    \caption{The comprehensive performance evaluation of PINN and ATL-PINNs.}
    \centering
    \small
    \begin{threeparttable}
        \begin{tabular}{cc|c|cc|cc|cc|cc}
        \hline
        PDE                         & Name     & PINN    & \multicolumn{2}{|c|}{Hard} & \multicolumn{2}{c|}{Soft} & \multicolumn{2}{c|}{MMoE} & \multicolumn{2}{c}{PLE} \\ \cline{3-11}
        -                           & -        & -       & \emph{Org}         & \emph{Cos}        & \emph{Org}         & \emph{Cos}        & \emph{Org}         & \emph{Cos}        & \emph{Org}        & \emph{Cos}        \\ \hline
        \multirow{3}{*}{Diff-React}   & \emph{L2 error} & 9.66E-2 & 3.87E-2     & 3.72E-2    & 4.01E-2     &   3.86E-2  &   3.62E-2   &   \textbf{3.62E-2}   & 3.91E-2   & 3.91E-2 \\
                                    & \emph{Boost}    & -       & 59.93\%     & 61.54\%    & 58.46\%     & 60.07\%    & 62.52\%     & \textbf{62.58\%}    & 59.53\%    & 59.54\%    \\ 
                                    & \emph{Number}   & -       & 63/100      & 63/100     & 59/100      & 67/100     & 68/100      & \textbf{70/100}     & 63/100     & 60/100     \\ \hline
        \multirow{3}{*}{Burgers'}   & \emph{L2 error} & 1.49E-1 & 1.37E-1     & 1.32E-1    & 1.38E-1     & 1.33E-1    & 1.35E-1  & 1.35E-1   & 1.28E-1    & \textbf{1.27E-1}    \\
                                    & \emph{Boost}    & -       & 7.92\%	   & 11.38\%	& 7.00\%	  & 10.31\%	   & 9.37\% 
        & 9.18\%	& 13.60\%	 & \textbf{14.30\%} \\
                                    & \emph{Number}   & -       & 84/100      & \textbf{90/100}     & 76/100      & 86/100     & 86/100      & 82/100     & 81/100     & 85/100     \\ \hline
        \multirow{3}{*}{Sha-Water}  & \emph{L2 error} & 2.73E-2 & 2.33E-2     & \textbf{2.24E-2}    & 2.33E-2     & 2.28E-2    & 2.37E-2     & 2.37E-2    & 2.44E-2    & 2.45E-2    \\
                                    & \emph{Boost}    & -       & 14.67\%     & \textbf{17.91\%}    & 14.62\%     & 16.17\%    & 13.13\%     & 13.18\%    & 10.39\%    & 10.24\%    \\
                                    & \emph{Number}   & -       & 61/100      & 70/100     & 73/100      & \textbf{77/100}     & 68/100      & 69/100     & 64/100     & 61/100   \\  \hline
        \end{tabular}
        \begin{tablenotes}    
            \footnotesize              
            \item[1] \emph{Org} denote to the original version and \emph{Cos} denote to the gradient cosine similarity version. 
            \item[2] \emph{L2 error} represents the average L2-related error across the 100 tasks under study. 
            \item[3] \emph{Boost} indicates the average percentage of improvement in prediction accuracy achieved by each mode. 
            \item[4] \emph{Number} quantifies the number of tasks in which the corresponding mode outperforms the PINN baseline.       
        \end{tablenotes}
    \end{threeparttable}
    \label{tab1}
\end{table}

\subsection{Diffusion Reaction equation}

The one-dimensional time-space Diffusion Reaction equation and its corresponding initial conditions are illustrated by the following equations:
\begin{gather}
    \partial_tu\left(t,x\right)-\upsilon\partial_{xx}\left(t,x\right)-\rho u\left(1-u\right)=0,\ x\in(0,1),\ t\in(0,1] \label{eq10} \\
    x\in(0,1),\ t\in(0,1] \label{eq11}
\end{gather}

This equation presents a challenging problem that combines a rapid evolution from a source term with a diffusion process, thus testing the network's capability to capture swift dynamics accurately. Here, the diffusion coefficient is represented by $\nu=0.5$, and the mass density is denoted by $\rho=1$. The boundary condition is periodic, and the initial condition, described as \eqref{eq12}, consists of a superposition of sinusoidal waves:

\begin{equation}
    u_0\left(x\right)=\sum_{k_i=k_1,...,k_N}{A_i\sin{{(2\pi{n_i}/L_x)x+\emptyset_i}}} \label{eq12}
\end{equation}

where $L$ represents the size of the calculation domain; $n_i$, $A_i$, and $\varphi_i$ denote random sample values. The value of $n_i$ is a random integer within the range $[1, 8]$, $A_i$ is a uniformly chosen random float number between 0 and 1, and $\varphi_i$ is a randomly selected phase within the interval $(0, 2\pi)$. We note that N=2 in this equation. The spatiotemporal domain used for training ranges from 0 to 1 in both spatial and temporal dimensions. We discretize them into $N_x\times N_t=1024\times 256$ points. To enforce the prediction that satisfied the initial condition and boundary conditions, we randomly choose 100 initial points and 100 boundary points in each boundary. The network learns the governing equation using a total of 10,000 sample points.

The network structure for single-task learning consisted of four layers, each with 100 cells. We use three layers, 100 cells each layer for expert networks and two layers, 100 cells each layer for tower networks in the auxiliary-task learning networks. All modes underwent training for 30,000 iterations using the Adam optimizer, with a learning rate of 10E-3 and the activation function set to $tanh$. Additionally, we implemente a learning rate decay method for the optimizer, reducing it to half of the original value every 10,000 iterations. It is important to note that unless explicitly stated, the training environment and network configuration remained consistent across all modes to ensure a fair comparison of their solution prediction accuracy.

\begin{figure}[htbp]
  \centering
  \includegraphics[width=0.9\linewidth]{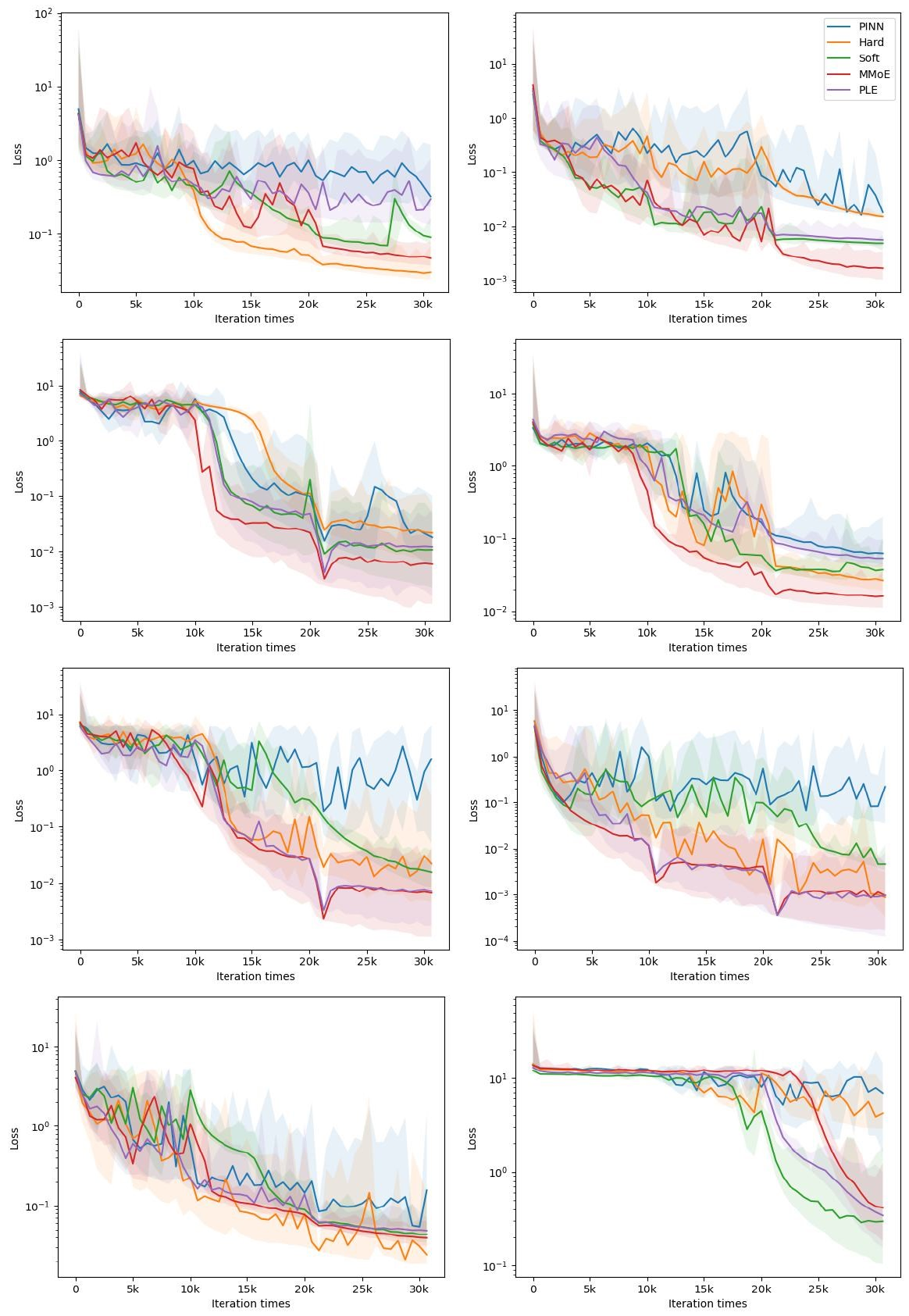}
  \caption{The convergence of PINN and ATL-PINNs (on the log scale) on the Diffusion Reaction equation in some example case studies.}
  \label{fig3}
\end{figure}

Table \ref{tab2} displays ten tasks exhibiting the most significant performance improvement across Diffusion Reaction equations in the experimental dataset. Nearly all the cases in the table achieve a performance enhancement of one order, with a maximum boost of approximately 96.62\%. Notably, the MMoE mode delivers outstanding results, accounting for half of the best-performing tasks. In Table \ref{tab2}, seven of the best results come from the \emph{Cos} version, demonstrating that the gradient cosine similarity algorithm can benefit auxiliary-task learning. Figure \ref{fig3} illustrates the convergence situation of the loss function for different cases and corresponding modes. We apply Gaussian smoothing to the loss curves, which enhances clarity and distinguishes between methods in a single figure. The blue lines represent the loss variation of PINN, and other colorful lines denote to loss variation of different modes in ATL-PINNs. We can see that the blue lines are nearly always higher than other lines, demonstrating that ATL-PINNs converge better than PINNs. In figure \ref{fig3}, the MMoE and PLE modes exhibit less pronounced fluctuations than the Hard and Soft modes, owing to the presence of the attention mechanism introduced by the gate networks. However, the final performance does not follow a consistent pattern, indicating that the network architecture can influence predictions across different tasks. Overall, the auxiliary-task approach yields an average improvement of 62.58\% (maximum improvement of 96.62\%) in the dataset comprising various diffusion equations, showing its immense potential.

\begin{table}[htbp]
    \caption{The case studies of best 10 boosting in Diffusion Reaction equation.}
    \centering
    \small
    \begin{tabular}{c|c|cc|cc|cc|cc|c}
        \hline
        \multirow{2}{*}{Subtask} & PINN    & \multicolumn{2}{c}{Hard} & \multicolumn{2}{c}{Soft} & \multicolumn{2}{c}{MMoE} & \multicolumn{2}{c|}{PLE} & \multirow{2}{*}{Max Boost} \\ \cline{2-10}
                                 & -       & \emph{Org}         & \emph{Cos}        & \emph{Org}         & \emph{Cos}        & \emph{Org}         & \emph{Cos}        & \emph{Org}        & \emph{Cos}        &                            \\
        \hline
        0                   & 7.19E-1 & 2.90E-2     & 2.97E-2    & 2.45E-2     & 2.51E-2    & 2.53E-2     & \textbf{2.43E-2}    & 2.90E-2    & 2.81E-2    & 96.62\%                    \\
        1                   & 6.93E-1 & 3.12E-2     & \textbf{2.80E-2}    & 3.41E-2     & 3.05E-2    & 2.98E-2     & 2.98E-2    & 3.05E-2    & 2.96E-2    & 95.96\%                    \\
        2                   & 5.24E-1 & 2.50E-2     & 2.78E-2    & 2.61E-2     & 2.90E-2    & \textbf{2.31E-2}     & 2.43E-2    & 3.42E-2    & 3.52E-2    & 95.59\%                    \\
        3                   & 7.45E-1 & 3.85E-2     & 4.41E-2    & 3.41E-2     & 3.91E-2    & 3.41E-2     & \textbf{3.31E-2}    & 4.79E-2    & 4.70E-2    & 95.56\%                    \\
        4                   & 6.92E-1 & 5.99E-2     & 5.15E-2    & 3.22E-2     & \textbf{2.76E-2}    & 4.45E-2     & 4.27E-2    & 3.22E-2    & 3.28E-2    & 95.35\%                    \\
        5                   & 6.47E-1 & 3.53E-2     & 3.12E-2    & 5.50E-2     & 4.86E-2    & \textbf{3.53E-2}     & 3.57E-2    & 6.02E-2    & 6.32E-2    & 95.17\%                    \\
        6                   & 6.67E-1 & 5.99E-2     & 5.62E-2    & 4.47E-2     & \textbf{4.19E-2}    & 4.47E-2     & 4.64E-2    & 6.80E-2    & 7.07E-2    & 93.72\%                    \\
        7                   & 1.67E-1 & \textbf{2.02E-2}     & 2.23E-2    & 2.97E-2     & 3.28E-2    & 3.11E-2     & 3.02E-2    & 3.18E-2    & 3.06E-2    & 87.92\%                    \\
        8                   & 2.21E-1 & 3.65E-2     & \textbf{3.12E-2}    & 3.81E-2     & 3.25E-2    & 3.37E-2     & 3.20E-2    & 3.94E-2    & 3.90E-2    & 85.91\%                    \\
        9                   & 1.84E-1 & 5.59E-2     & 3.82E-2    & 4.84E-2     & 3.31E-2    & 2.85E-2     & \textbf{2.70E-2}   & 2.81E-2    & 2.81E-2    & 85.34\%    \\
        \hline
    \end{tabular}
    \label{tab2}
\end{table}

\subsection{Burger’s equation}

The one-dimensional time-space Burger’s equation, along with its corresponding initial conditions, represents a mathematical mode for capturing the nonlinear behavior and diffusion process in fluid dynamics. Specifically, the equation and initial conditions are defined as follows:
\begin{gather}
    \partial_tu\left(t,x\right)+\partial_x\left(u^2(t,x\right)/2)=\nu/\pi\partial_{xx}u\left(t,x\right),\ x\in(0,1),\ t\in(0,2] \label{eq13} \\
    u\left(0,x\right)=u_0\left(x\right),\ x\in(0,1) \label{eq14}
\end{gather}
where $\nu = 0.01$ represents the diffusion coefficient. The boundary condition is periodic and the initial condition is described in \eqref{eq12}.

In this case study, the networks are trained on a spatiotemporal domain of $[0,1]\times[0,2]$, discretized into $N_x\times N_t=1024\times 256$ points. The network structure for single-task learning consisted of five layers, each with 50 cells. For the auxiliary-task learning networks, we utilize four layers with 50 cells each for the shared experts and two layers with 50 cells each for the tower networks. All other settings remained the same as in the Diffusion Reaction equation (e.g., number of points, optimizer configuration).

\begin{figure}[htbp]
  \centering
  \includegraphics[width=\linewidth]{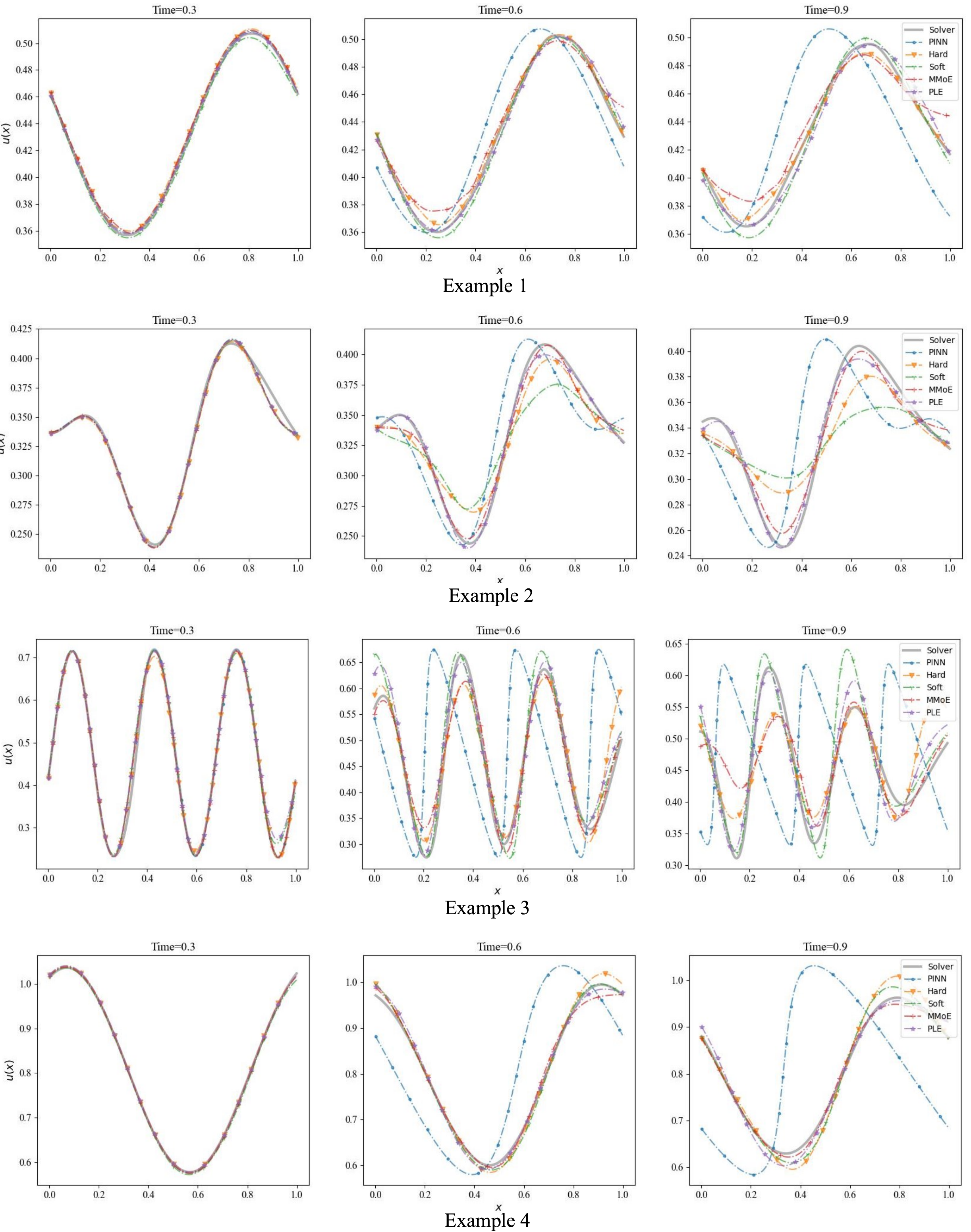}
  \caption{The prediction of PINN and ATL-PINNs at three different times ($t=0.3$, $t=0.6$, and $t=1$) on the Burgers' equation in some case studies.}
  \label{fig4}
\end{figure}

Table \ref{tab3} shows notable improvements achieve through auxiliary-task learning. While the maximum boost efficiency may not be as high as observed in the Diffusion Reaction equation, the enhancement range is significantly broader, with improvements seen in nearly ninety percentage cases. Figure \ref{fig4} presents the predictions of various modes at three different time points. The gray lines represent the reference solution obtained from the solver, the blue lines depict the forecasts from the PINN baseline, and the other lines correspond to the auxiliary-task modes. In Figure \ref{fig4}, both the PINN and ATL-PINNs initially fit the solution well. However, as time progresses and the equation evolves, the predictions from the PINN gradually deviate from the reference. At $t=0.6$, PINN starts to depart from the reference solution while ATL-PINNs can still predict the result with litter error in their peaks and troughs. At $t=0.9$, the PINN fails to capture the waveform accurately. However, the ATL-PINNs perform significantly better. Although deviations are present in their peaks and troughs, the waveforms are generally correct from ATL-PINNs' prediction. In Figure \ref{fig4}, the PLE modes can get the best average result, shown as purple lines, which fit the reference most precisely. Table \ref{tab1} demonstrates this result. Overall, the ATL-PINNs yield an average improvement of 14.30\% (maximum improvement of 36.24\%) over ninety percent of cases in the Burgers' equation dataset, revealing the universality of using auxiliary-task learning method to improve PINNs for PDE solving.

\begin{table}[htbp]
    \caption{The case studies of best 10 boosting in Burgers' equation.}
    \centering
    \small
    \begin{tabular}{c|c|cc|cc|cc|cc|c}
        \hline
        \multirow{2}{*}{Subtask} & PINN    & \multicolumn{2}{c}{Hard} & \multicolumn{2}{c}{Soft} & \multicolumn{2}{c}{MMoE} & \multicolumn{2}{c|}{PLE} & \multirow{2}{*}{Max Boost} \\ \cline{2-10}
                                 & -       & \emph{Org}         & \emph{Cos}        & \emph{Org}         & \emph{Cos}        & \emph{Org}         & \emph{Cos}        & \emph{Org}        & \emph{Cos}        &    \\
        \hline
        0                   & 2.04E-1 & 1.33E-1     & 1.33E-1    & 1.33E-1     & 1.33E-1    & \textbf{1.30E-1}     & 1.30E-1    & 1.69E-1    & 1.69E-1    & 36.24\%                    \\
        1                   & 1.61E-1 & 1.41E-1     & 1.24E-1    & 1.18E-1     & \textbf{1.04E-1}    & 1.20E-1     & 1.20E-1    & 1.35E-1    & 1.28E-1    & 35.44\%                    \\
        2                   & 2.07E-1 & 1.66E-1     & 1.53E-1    & 1.66E-1     & 1.53E-1    & 1.53E-1     & 1.45E-1    & 1.47E-1    & \textbf{1.40E-1}    & 32.39\%                    \\
        3                   & 1.06E-1 & 8.90E-2     & 8.84E-2    & 7.72E-2     & 7.67E-2    & \textbf{7.25E-2}     & 7.61E-2    & 8.72E-2    & 8.72E-2    & 31.60\%                    \\
        4                   & 1.51E-1 & \textbf{1.05E-1}     & 1.06E-1    & 1.27E-1     & 1.29E-1    & 1.12E-1     & 1.18E-1    & 1.25E-1    & 1.31E-1    & 30.61\%                    \\
        5                   & 1.42E-1 & 1.04E-1     & \textbf{9.92E-2}    & 1.26E-1     & 1.20E-1    & 1.08E-1     & 1.03E-1    & 1.03E-1    & 1.03E-1    & 30.22\%                    \\
        6                   & 1.79E-1 & 1.55E-1     & 1.48E-1    & 1.31E-1     & \textbf{1.26E-1}    & 1.52E-1     & 1.60E-1    & 1.48E-1    & 1.41E-1    & 29.77\%                    \\
        7                   & 1.98E-1 & 1.55E-1     & 1.59E-1    & \textbf{1.42E-1}     & 1.46E-1    & 1.50E-1     & 1.50E-1    & 2.03E-1    & 2.03E-1    & 28.12\%                    \\
        8                   & 1.88E-1 & 1.41E-1     & \textbf{1.38E-1}    & 1.48E-1     & 1.44E-1    & 1.42E-1     & 1.49E-1    & 1.65E-1    & 1.73E-1    & 26.98\%                    \\
        9                   & 1.36E-1 & 1.08E-1     & 1.07E-1    & 1.17E-1     & 1.16E-1    & \textbf{1.03E-1}     & 1.03E-1    & 1.16E-1    & 1.16E-1    & 24.04\%        \\           
         \hline
    \end{tabular}
    \label{tab3}
\end{table}

\subsection{Shallow Water equations}

The two-dimensional time-space Shallow Water equations, denoted as equations \eqref{eq15} to \eqref{eq17}, capture the dynamics of free-surface flow problems:
\begin{gather}
    \partial_th+\partial_xhu+\partial_yhu=0 \label{eq15} \\
    \partial_thu+\partial_x\left(u^2h+\frac{1}{2}g_rh^2\right)=-g_rh\partial_xb \label{eq16} \\
    \partial_thv+\partial_y\left(v^2h+\frac{1}{2}g_rh^2\right)=-g_rh\partial_yb \label{eq17} 
\end{gather}
where the variables $u$ and $v$ represent the velocities in the horizontal and vertical directions, respectively. h corresponds to the water depth, which serves as the primary prediction in this problem. Additionally, $g_r=1.0$ represent the gravitational acceleration. Derived from the general Navier-Stokes (N-S) equations, these equations find broad application in modeing various free-surface flow phenomena.

The dataset provided by PDEBench shows a 2D radial dam break scenario that captures the evolution process of a circular bump within a square domain. This scenario involves initializing the water height at the center. The initial condition for $h$ can be defined as follows:
\begin{equation}
    h=\left\{\begin{matrix}2.0,&for\ r\ <\sqrt{x^2+y^2}\\1.0,&for\ r\ \geq\sqrt{x^2+y^2}\\\end{matrix}\right. \label{eq18}
\end{equation}
where the radius $r$ randomly sampled from $U(0.3,0.7)$. The spatial dimension for training is $\Omega=[-2.5,2.5]\times[-2.5,2.5]$; the temporal dimension spans $T=[0,1]$. To discretize the dataset, we employ a resolution of $N_x\times N_y\times N_t=128\times 128\times 101$.

In contrast to the one-dimensional case studies, the two-dimensional problem presents greater complexity. As a result, we employ a deeper network with an increased number of neural cells to address these equations. The network structure for single-task learning consists of six layers, each comprising 100 cells. In the auxiliary-task learning networks, we utilize five layers with 100 cells each for the expert networks and two layers with 100 cells each for the tower networks. To accommodate the increased complexity, the number of points is also augmented. Specifically, we employ 1000 boundary points for each boundary, 1000 initial points, and 100,000 intra-domain sample points to train both networks. Because the number of sample points is too large to input into the network once, we employ a mini-batch approach by randomly selecting 20,000 sample points for each iteration. It is important to note that all the boundary and initial points are sent to the network during every iteration. As for other network configurations, such as the optimizer, we keep them the same as those employed in the burger’s equation experiment.

\begin{figure}[]
  \centering
  \includegraphics[width=0.95\linewidth]{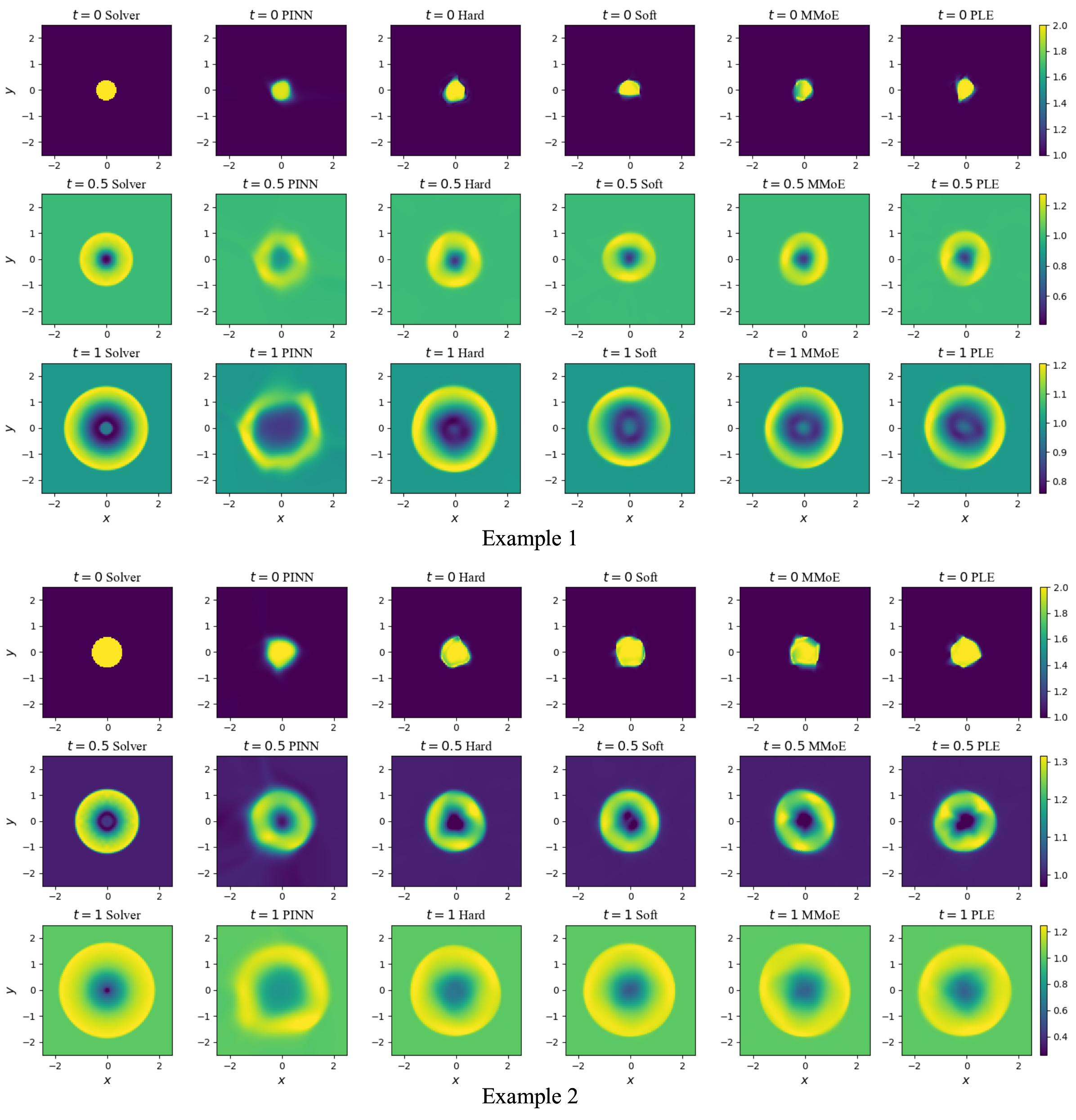}
  \caption{The reference solution and prediction of PINN and ATL-PINNs at three different times ($t=0.0$, $t=0.5$, and $t=1$) on the Shallow Water equation in some case studies.}
  \label{fig5}
\end{figure}

Table \ref{tab4} presents the top ten cases where auxiliary-task learning yields the best benefits in the Shallow Water equation. The ATL-PINNs can achieve a maximum boost of approximately 72\%. In Table \ref{tab4}, the Hard mode consistently exhibits the most substantial enhancement among the different modes. We get an unexpected result that the most straightforward modes achieve the best results in the most complex problems. One reason behind this phenomenon is perhaps the relevance between different tasks is strong (only the initial water height is changed). Therefore, the Hard model can learn the shared representation more directly, thus leading to better results. Figure \ref{fig5} shows the predictions of the water height ($h$) in two example cases at three-time points ($t=0$, $t=0.5$, and $t=1$). The Shallow Water equation poses a challenging problem for physics-informed learning due to its 2D complexity and large solution domain. Therefore, the PINN baseline struggles to accurately capture the circle form of water height in the center dam. The results in the ATL-PINNs deliver superior results. Although the water height cannot be prediction accuracy, the ATL-PINN can portray the circle form of the water height in the center dam. All the ATL-PINN modes can better capture the underlying physics processes and improve prediction accuracy by incorporating auxiliary knowledge from tasks. Overall, the auxiliary-task approach yields an average improvement of 17.91\% (maximum improvement of 72.38\%) in the Shallow Water equation dataset, revealing the potential for handling complex scenarios in physics problems.

\begin{table}[htbp]
    \caption{The detail of the best 10 boosting case studies in Shallow Water equation.}
    \centering
    \small
    \begin{tabular}{c|c|cc|cc|cc|cc|c}
        \hline
        \multirow{2}{*}{Subtask} & PINN    & \multicolumn{2}{c}{Hard} & \multicolumn{2}{c}{Soft} & \multicolumn{2}{c}{MMoE} & \multicolumn{2}{c|}{PLE} & \multirow{2}{*}{Max Boost} \\ \cline{2-10}
                                 & -       & \emph{Org}         & \emph{Cos}        & \emph{Org}         & \emph{Cos}        & \emph{Org}         & \emph{Cos}        & \emph{Org}        & \emph{Cos}        &    \\
        \hline
        0                   & 2.98E-2 & 1.88E-2     & 2.40E-2    & 2.20E-2     & 2.81E-2    & \textbf{1.71E-2}     & 1.80E-2    & 1.74E-2    & 1.74E-2    & 72.38\%                    \\
        1                   & 3.26E-2 & \textbf{1.66E-2}     & 2.31E-2    & 2.05E-2     & 2.86E-2    & 2.14E-2     & 2.14E-2    & 2.12E-2    & 2.23E-2    & 60.77\%                    \\
        2                   & 4.82E-2 & \textbf{1.96E-2}     & 2.46E-2    & 2.14E-2     & 2.69E-2    & 3.18E-2     & 3.18E-2    & 4.12E-2    & 3.91E-2    & 59.99\%                    \\
        3                   & 5.20E-2 & 3.35E-2     & 2.08E-2    & 1.52E-2     & \textbf{9.47E-3}    & 3.54E-2     & 3.71E-2    & 4.18E-2    & 4.18E-2    & 59.35\%                    \\
        4                   & 3.58E-2 & 2.46E-2     & \textbf{1.72E-2}    & 3.15E-2     & 2.21E-2    & 2.30E-2     & 2.18E-2    & 2.43E-2    & 2.43E-2    & 58.70\%                    \\
        5                   & 4.48E-2 & \textbf{2.33E-2}     & 2.43E-2    & 2.50E-2     & 2.60E-2    & 3.28E-2     & 3.28E-2    & 3.28E-2    & 3.12E-2    & 58.08\%                    \\
        6                   & 3.45E-2 & \textbf{1.43E-2}     & 1.76E-2    & 1.98E-2     & 2.44E-2    & 2.36E-2     & 2.47E-2    & 1.89E-2    & 1.98E-2    & 51.79\%                    \\
        7                   & 5.01E-2 & 3.63E-2     & \textbf{1.97E-2}    & 4.61E-2     & 2.50E-2    & 3.33E-2     & 3.49E-2    & 2.13E-2    & 2.03E-2    & 49.22\%                    \\
        8                   & 3.35E-2 & \textbf{1.41E-2}     & 1.57E-2    & 2.27E-2     & 2.54E-2    & 2.36E-2     & 2.36E-2    & 1.81E-2    & 1.81E-2    & 48.05\%                    \\
        9                   & 5.91E-2 & \textbf{1.63E-2}     & 2.80E-2    & 1.67E-2     & 2.86E-2    & 2.63E-2     & 2.76E-2    & 2.29E-2    & 2.29E-2    & 42.52\%      \\
        \hline
    \end{tabular}
    \label{tab4}
\end{table}

\section{Conclusion}
\label{Conclusion}

Motivated by the remarkable success of shared common representation in auxiliary-task learning, our study incorporates these modes into neural network-based surrogate models to investigate the potential benefits of PINNs through correlated auxiliary-task learning. Specifically, we randomly select auxiliary tasks with different initial conditions in homogeneous PDEs. In this paper, we propose ATL-PINNs with four modes for auxiliary-task learning and test them in three PDE datasets, each involving 100 tasks that can leverage the performance of the auxiliary-task learning modes. We also introduce the gradient cosine similarity approach to ensure that the updates from the auxiliary task consistently benefit the main problem. The experimental results demonstrate that the auxiliary-task learning modes enhance the performance of network-based surrogate models in physics-informed learning, and the gradient cosine similarity approach further improves their performance. However, selecting suitable auxiliary tasks and determining the optimal number for the main problem remains unexplored. In future work, we will focus on developing efficient algorithms for auxiliary-task construction and selection. We are also interested in exploring the application of auxiliary-task learning modes in more complex real-world scenarios. Although this paper represents a preliminary attempt to combine auxiliary-task learning and PINNs, we hope our research will contribute to the broader application of auxiliary-task learning-based modes in the physics-informed learning scene.

\section*{Acknowledgments}
This research work was supported by the National Key Research and Development Program of China (2021YFB0300101). The datasets and code used during the current study can be accessed on GitHub: https://github.com/junjun-yan/ATL-PINN. The authors declare no conflict of interest.

\bibliographystyle{unsrt}  
\bibliography{references}

\end{document}